\documentclass{article}

\usepackage{arxiv}

\usepackage[utf8]{inputenc} % allow utf-8 input
\usepackage[T1]{fontenc}    % use 8-bit T1 fonts
\usepackage{hyperref}       % hyperlinks
\usepackage{url}            % simple URL typesetting
\usepackage{booktabs}       % professional-quality tables
\usepackage{amsfonts}       % blackboard math symbols
\usepackage{nicefrac}       % compact symbols for 1/2, etc.
\usepackage{microtype}      % microtypography
\usepackage{lipsum}
\usepackage{graphicx}
\usepackage{float}

\RequirePackage{algorithm}
\RequirePackage{algorithmic}
\title{SAFE ML: Surrogate Assisted Feature Extraction for Model Learning}

\author{
  Alicja Gosiewska \\
  Faculty of Mathematics and Information Science\\
  Warsaw University of Technology\\
  \texttt{a.gosiewska@mini.pw.edu.pl} \\
   \And
 Aleksandra Gacek \\
   Faculty of Mathematics and Information Science\\
  Warsaw University of Technology\\
  \texttt{a.gacek@student.mini.pw.edu.pl} \\
  \AND
 Piotr Lubon \\
  Faculty of Mathematics and Information Science\\
  Warsaw University of Technology\\
  \texttt{lubonp@student.mini.pw.edu.pl} \\
   \And
 Przemyslaw Biecek \\
    Faculty of Mathematics, Informatics and Mechanics\\ 
    University of Warsaw \\
  Faculty of Mathematics and Information Science\\
  Warsaw University of Technology\\
  \texttt{przemyslaw.biecek@gmail.com} \\
}

\begin{document}
\maketitle

\begin{abstract}
Complex black-box predictive models may have high accuracy, but opacity causes problems like lack of trust, lack of stability, sensitivity to concept drift. On the other hand, interpretable models require more work related to feature engineering, which is very time consuming. Can we train interpretable and accurate models, without timeless feature engineering? 
In this article, we show a~method that uses elastic black-boxes as surrogate models to create a simpler, less opaque, yet still accurate and interpretable glass-box models. New models are created on newly engineered features extracted/learned with the help of a~surrogate model.
We show applications of this method for model level explanations and possible extensions for instance level explanations. 
We also present an example implementation in Python and benchmark this method on a number of tabular data sets.
\end{abstract}

\section{Motivation}
\label{motivation}

Questions of trust in machine learning models became crucial issues in recent years. Complex predictive models have various applications in different areas \cite{PALIWAL20092, KOUROU20158} and an increasing number of people use machine learning solutions in everyday life. Hence, it is important to ensure that predictions of these models are reliable. There are four requirements whose fulfillment is essential to ensure that predictive model is trustworthy and accessible: (1) high model performance, (2) auditability, (3) interpretability, and (4) automaticity.

(1) High model performance means that a model rarely makes wrong predictions or the prediction error is small on average. Usually, this can be achieved by using complex, so-called black-box models, such as, boosting trees \cite{DBLP:journals/corr/ChenG16} or deep neutral networks \cite{Goodfellow-et-al-2016}. The opposite of black-boxes are glass-boxes. They are simple, interpretable models, such as linear regression, logistic regression, decision trees, regression trees, and decision rules.

Model performance ensures only a part of information about model's quality. Model's (2) auditability guarantees that the model can be verified with respect to different criteria. They are, for example, stability, fairness, and sensitivity to a~concept drift. There are tools that allow to audit black-box models \cite{gosiewska2018auditor}, yet simple glass-boxes offer more extended range of diagnostic methods \cite{Harrell:2006:RMS:1196963}.

The third requirement is an (3) interpretability, which became an important topic in recent years \cite{ONeil}. Machine learning models influence people's lives, in particular, they are used by financial, medical, and security institutions. Models have an impact on whether we get a~loan \cite{HUANG2007847}, what type of treatment we receive \cite{doi:10.1177/117693510600200030}, or even whether we are searched by the police \cite{4053200}. Therefore, models reasoning should be transparent and accessible. There is an ongoing debate about the right to explanation, what does it mean and how it can be achieved \cite{DBLP:journals/corr/abs-1711-00399, Edwards_Veale_2018}.

The (4) automaticity of machine learning methods is spreading rapidly.
Due to the increasing computational power, it becomes easier and easier to obtain more precise models, usually in an automatic manner. There are automated frameworks for AutoML like autokeras, auto-sklearn, TPOT \cite{jin2018efficient, NIPS2015_5872, Olson2016} that allow one to train a model even without any statistical knowledge or even programming skills. Yet, machine learning specialists can also take an advantage of automated methods of modeling. Such methods reduce time needed to train the model, therefore human effort can be directed towards more creative and sophisticated tasks than testing wide range of parameters and models.

People usually choose automatically fitted black-box models that achieve high performance at the cost of auditability and interpretability. 
As a response to this problem, the methodology for explaining predictions of black-box models, so called post-hoc interpretability, is under active development. There are several approaches to explaining the global behavior of black-boxes. Model can be reduced to simple if-then rules \cite{MAGIX} or decision trees \cite{proc-jsm-2018}. 
However, these explanations are simplifications of models and may be inaccurate. As a consequence, they may be misleading or even harmful. Hence, in many applications it is better to train a transparent, interpretable model than apply explanations to a complex model \cite{2017arXiv171006169T, pleseStop}. Therefore, automated methods of obtaining interpretable models, while maintaining the predictive capabilities of a~complex model, are extremely important.

In this article, we present a method for Surrogate Assisted Feature Extraction for Model Learning (SAFE ML). This method uses a surrogate model to assist feature engineering and lead to training accurate and transparent glass-box model. In this approach, surrogate model should be accurate to produce best feature transformation, yet it does not have to be interpretable. Based on the new features, the transparent glass-box model is trained. In many cases the high accuracy of black-box models comes from good data representation and this is something than can be next extracted from the model.

The SAFE ML method is flexible and model agnostic, any class of models may be used as a surrogate model and as a~glass-box model. Therefore, surrogate model may be selected to fit the data as best as possible, while glass-box model one can be selected according to the particular task or abilities of the end-users to interpreting models. 

An advantage of this methodology is that the final glass-box model has a performance close to the surrogate model. By changing the representation of the data, SAFE ML allows to gain interpretability with minimal or no reduction of model~performance. 

The SAFE ML method can be used as a step in training a model with AutoML methods. We can use AutoML to fit elastic and complex model, then use SAFE to obtain a~transparent model.

The paper is organized as follows. 
Section~\ref{SAFE_algorithm} provides a~description of the SAFE algorithm. Section~\ref{SAFE_application} contains illustrations and benchmarks for the SAFE method for regression and classification problems. Extensions for instance-level approaches and interactions are discussed in Section~\ref{extension}.
Conclusions are in Section~\ref{discussion}.

\section{Description of the SAFE Algorithm}
\label{SAFE_algorithm}

\begin{figure*}[t!h]
    \centering
    \includegraphics[width=\textwidth]{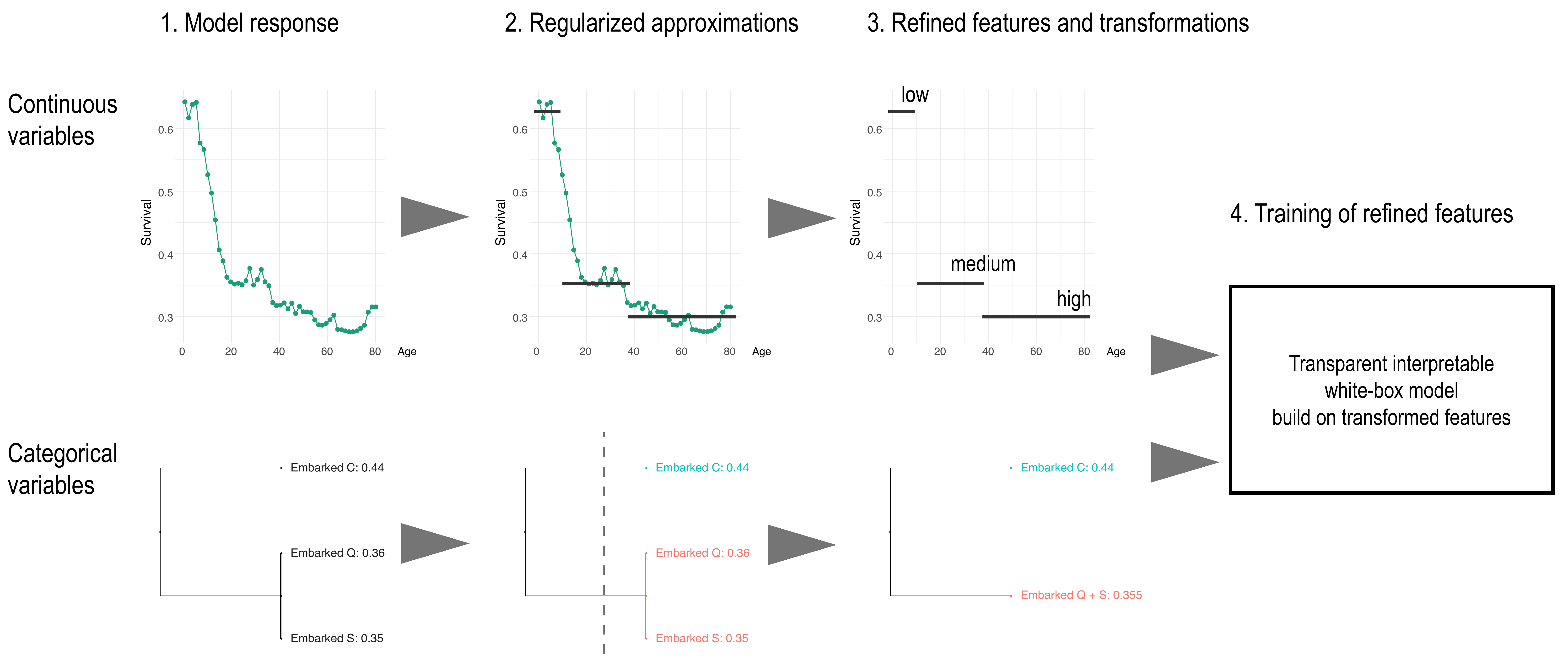}
    \caption{The SAFE ML algorithm in four steps, 1. train elastic surrogate model, 2. approximate model response, 3.~extract transformations and new features, 4. train refined model.}
    \label{fig:safeDiagram}
\end{figure*}

The SAFE ML algorithm uses a complex model as a surrogate. New binary features are created on the basis of surrogate predictions. These new features are used to train a simple refined model.

Illustration of the SAFE ML method is presented in Figure~\ref{fig:safeDiagram}.
In the Algorithm~\ref{alg:SAFEdescription} we describe how data transformations are extracted from the surrogate model while in Algorithm~\ref{alg:SAFElearning} we show how to train a new refined model based on transformed features. Below, we explain details of the terminology being used in algorithms.
Let $x_1, x_2, ..., x_p$ be features in the surrogate model $M$. A~subset of all features except $x_i$ we denote as $x_{-i}$.

\textbf{The partial dependence profile} \cite{PDP} is defined as
$$
f_i(x_i) = \mathbb{E}_{x_{-i}}[ M(x_i, x_{-i}) ],
$$
and calculated as
$$
\hat f_i(x_i) = \frac{1}{n} \sum_{j=1}^{n} M(x_{i}^j, x_{-i}^j),
$$
where $n$ is the number of observations and  $x_i^j$ is a value of the $i$-th feature for the $j$-th instance. Partial dependence function describes the expected output condition on a selected variable. The visualization of this function is Partial Dependence Plot \cite{RJ-2017-016}, an example plot is presented in Step~1 in Figure~\ref{fig:safeDiagram}.

\textbf{The change point method} \cite{DBLP:journals/corr/abs-1801-00718} is used to identify times when the probability distribution of a time series changes.
\textbf{The hierarchical clustering} \cite{Rokach2005} is an algorithm that groups observations into clusters. It involves creating a hierarchy of clusters that have a predetermined ordering. 
Step~2 in \mbox{Figure~\ref{fig:safeDiagram}} corresponds to both change point method and hierarchical clustering.

\begin{algorithm}[tb]
   \caption{Surrogate Assisted Feature Extraction}
   \label{alg:SAFEdescription}
\begin{algorithmic}
   \STATE {\bfseries Input:} data $X_{n \times p}$, surrogate model $M$, regularization penalty $\lambda$.
    \STATE {\bfseries Start:} 
   \FOR{$i=1$ {\bfseries to} $p$} 
   \STATE Let $x_i$ be $i$-th feature.
       \IF{$x_i$ is numerical}
       \STATE Calculate partial dependence profile $f_i(x)$ for feature $x_i$.
       \STATE Approximate $f_i(x)$ with interpretable features $x^*_{i}$, for example, use the change point method to discretize the variable with regularization penalty~$\lambda_i$.
       \STATE Save transformation $t_i(x)$ that transforms $x_i$ into~$x_i^*$.
       \ENDIF
       \IF{$x_i$ is categorical}
       \STATE Calculate model responses for each observation with imputed each possible value of $x_i$.
       \STATE Merge levels of $f_i(x)$ with similar model responses, for example use the hierarchical clustering with number of clusters $\lambda_i$.
       \STATE Save transformation $t_i(x)$ that transforms $x_i$ into~$x_i^*$.
   \ENDIF
  \ENDFOR
 \STATE Sets of transformations $T^* = \{t_1, ..., t_p\}$ may be used to create new data $X^*$ from features $x_i^* = t_i(x_i)$.
\end{algorithmic}
\end{algorithm}

\begin{algorithm}[th]
   \caption{Model Learning with Surrogate Assisted Feature Extraction}
   \label{alg:SAFElearning}
\begin{algorithmic}
   \STATE {\bfseries Input:} data $X^{new}_{m \times p}$, set of transformations $T^*$ derived from surrogate model $M$.
     \STATE {\bfseries Start:} 
\STATE Transform dataset $X$ into $X^{*, new} = T^*(X^{new})$.
\STATE Create transparent model $M^{new}$ based on $X^{*, new}$.
\end{algorithmic}
\end{algorithm}

\section{Application and Benchmarks}
\label{SAFE_application}

In this section, we perform SAFE ML on selected data sets for regression and classification problems. A summary discussion of the results is conducted at the end of this section.

Examples are generated with scikit-learn models \cite{scikit-learn} and SafeTransformer. SafeTransformer is a~Python library that implements SAFE ML method. 

Code that generates artificial data sets and performs SAFE ML method and can be found in the GitHub repository: \url{https://github.com/agosiewska/SAFE_examples}.

\subsection{Classification - Artificial Data Set}
\label{subsection_classification_artificial}

We compare performance of naïve logistic regression, surrogate xgboost, and refined logistic regression. Here naïve regression means that we fill vanilla regression model without any feature engineering. 

This example is performed on the artificial data set SIMULD2 for binary classification.  SIMULD2 consists of 500 observations and three variables. Variable $y$ is a binary target. Variable $X1$ is continuous, uniform distributed at range from $-5$ to $5$ with normally distributed noise. Variable $X2$ is categorical with 40 levels.

As can be seen in Table~\ref{tab:class_results}, refined logistic regression performs better than the other two models. Refined logistic regression achieves even better accuracy and AUC than xgboost model, while being a more transparent model. 

It may be surprising that the refined model is better than the surrogate one, however there are some reasons for that. Elastic models are better to capture non-linear relations but at the price of larger variance for parameter estimation. In some cases the refined models will work on better features and will have less parameters to train, thus it can outperform the surrogate model.

\begin{table}[h!tb]
\caption{Results of the SAFE method for models trained on the SIMULD2 data set. SAFE was performed with penalty equals~$0.42625$.}
\label{tab:class_results}
\vskip 0.15in
\begin{center}
\begin{small}
\begin{sc}
\begin{tabular}{lcccr}
\toprule
Model & Accuracy & AUC  \\
\midrule
Naïve logistic regression        & 0.736 & 0.897  \\
Surrogate xgboost  & 0.960 & 0.982  \\
Refined logistic regression    & \textbf{0.976} & \textbf{0.989} \\
\bottomrule
\end{tabular}
\end{sc}
\end{small}
\end{center}
\vskip -0.1in
\end{table}

\begin{figure}[tb]
    \vskip 0.2in
    \begin{center}
    \centerline{\includegraphics[width=0.6\textwidth]{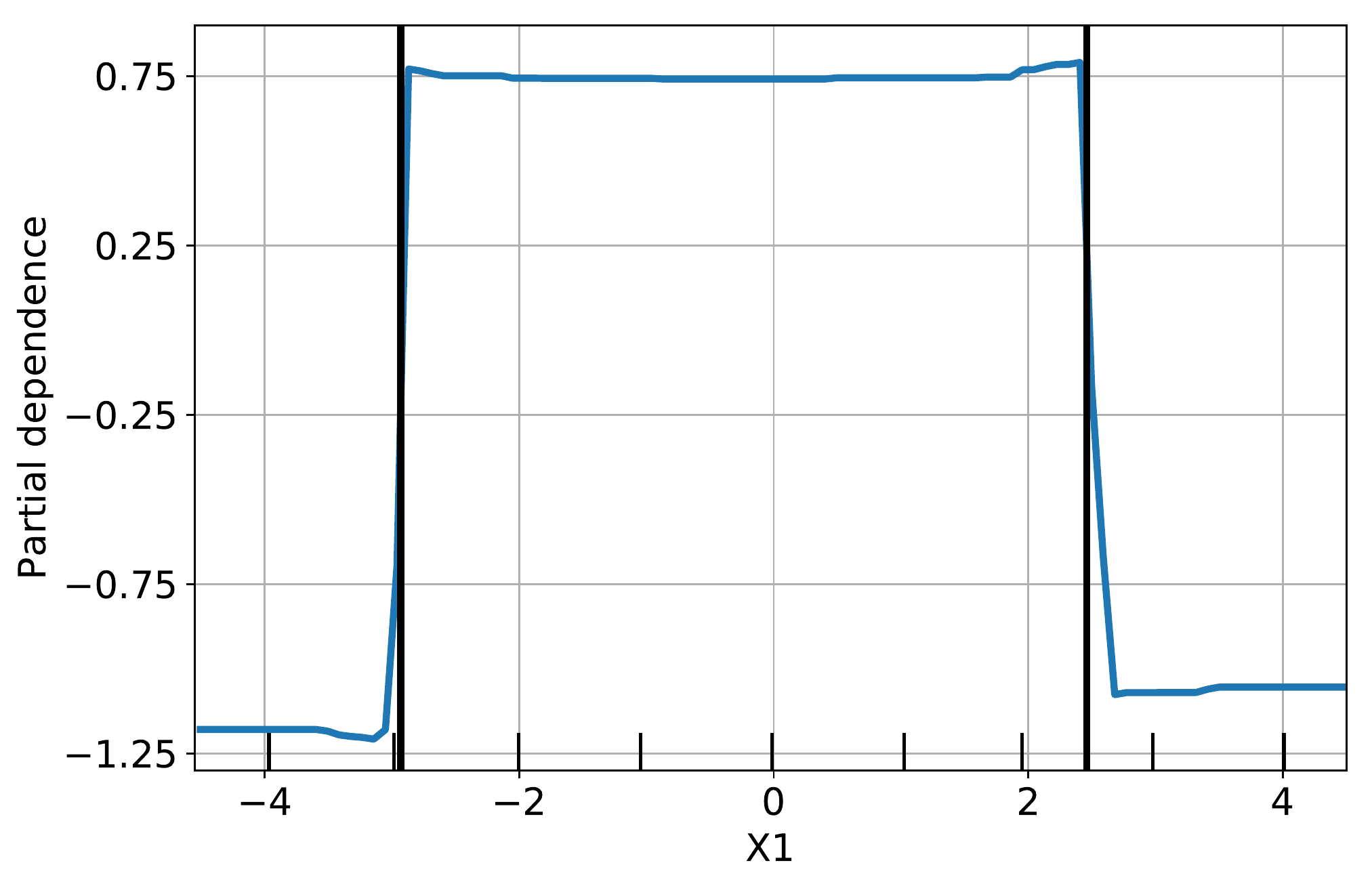}}
    \caption{An expected response of the \mbox{xgboost} model conditioned on the variable X1. Black vertical lines marks points of the discretization calculated with SAFE ML.}
    \label{fig:class_pdp}
    \end{center}
    \vskip -0.2in
\end{figure}

Partial Dependence Plot in Figure~\ref{fig:class_pdp}~shows the relationship between variable $X1$ and output of the xgboost model. This pattern is close to real association, which is a step function with discontinuities in $-3$ and $2.5$.
This relationship could not be caught by logistic regression. 
However, in Figure~\ref{fig:class_pdp},~we can see that SAFE ML method divided $X1$ variable into three binary variables. This make it possible for refined logistic regression to capture the non-linearity.

Variable $X2$ consists of 40 levels, yet process of generating target variable y distinguishes between variables in three groups. When examining how SAFE ML has grouped variables, one can see that groups almost match up with real dependencies. This caused that instead of one variable of 40 levels, the new model was trained on 3 binary variables.  This means that transformed features better reflected the real relationships.

\subsection{Regression - Boston Housing}
\label{subsection_regression_boston}

Second example is performed on Boston Housing data set \cite{HARRISON197881}. Boston Housing consists of 506 rows and 14 columns. The target variable is medv (median value of owner-occupied homes).

We compare performances of naïve linear regression, surrogate xgboost, and refined linear regression.  

\begin{figure}[tb]
    \vskip 0.2in
    \begin{center}
    \centerline{\includegraphics[width=0.6\textwidth]{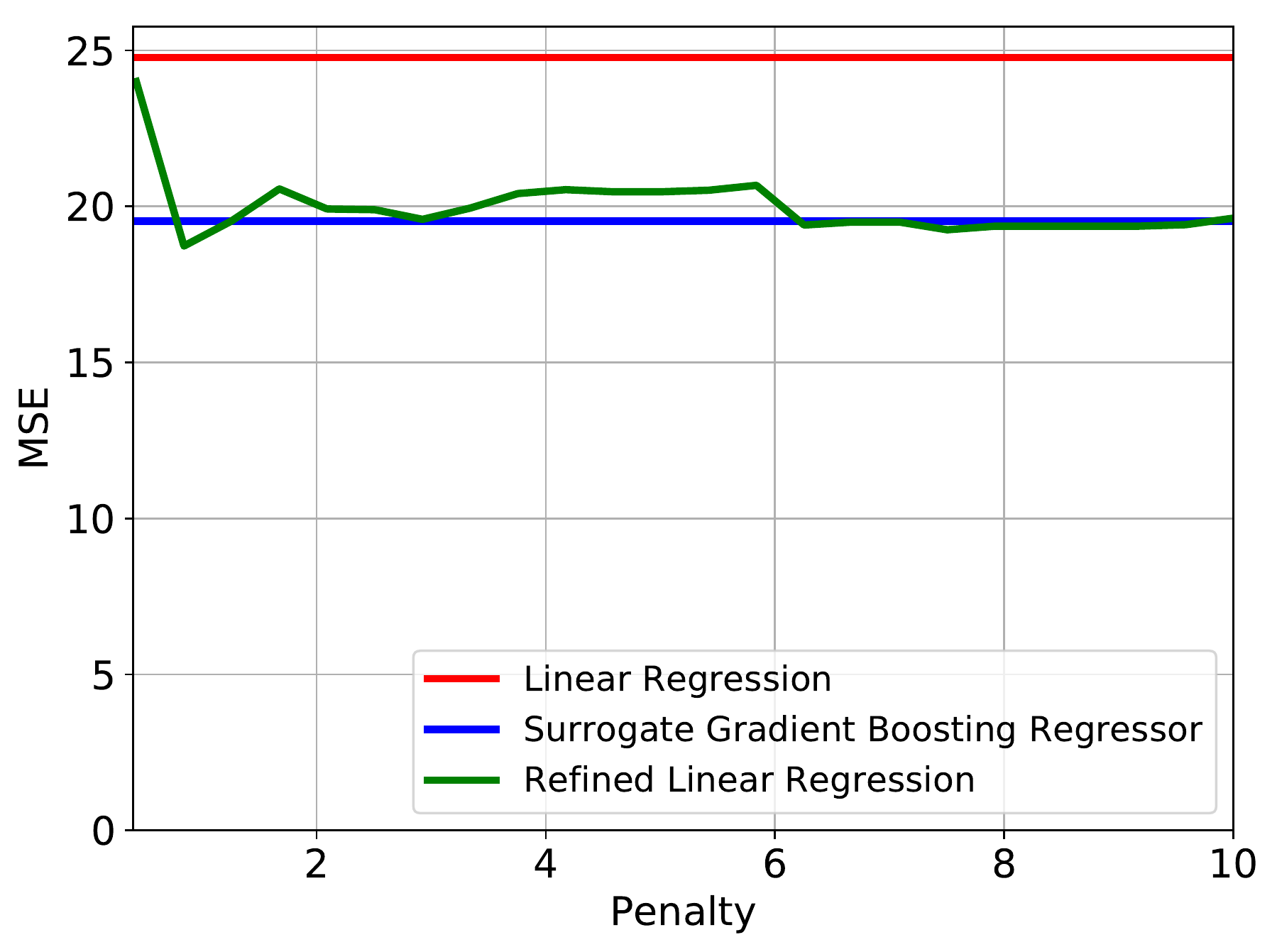}}
    \caption{Dependence between SAFE ML method's penalty and MSE for refined model. }
    \label{boston_results}
    \end{center}
    \vskip -0.2in
\end{figure}

As described in Section~\ref{SAFE_algorithm}, feature extraction in SAFE ML algorithm depends on a choice of a regularization penalty $\lambda$. Figure~\ref{boston_results} shows performances of models as functions of penalty. Mean Square Errors (MSE) of the refined linear regression models are, in general, close to MSE of surrogate model. Thus, the use of a simpler model did not negatively affect the performance. At the same time, we gained transparency.

\begin{figure}[tb]
\vskip 0.2in
\begin{center}
\centerline{\includegraphics[width=0.6\textwidth]{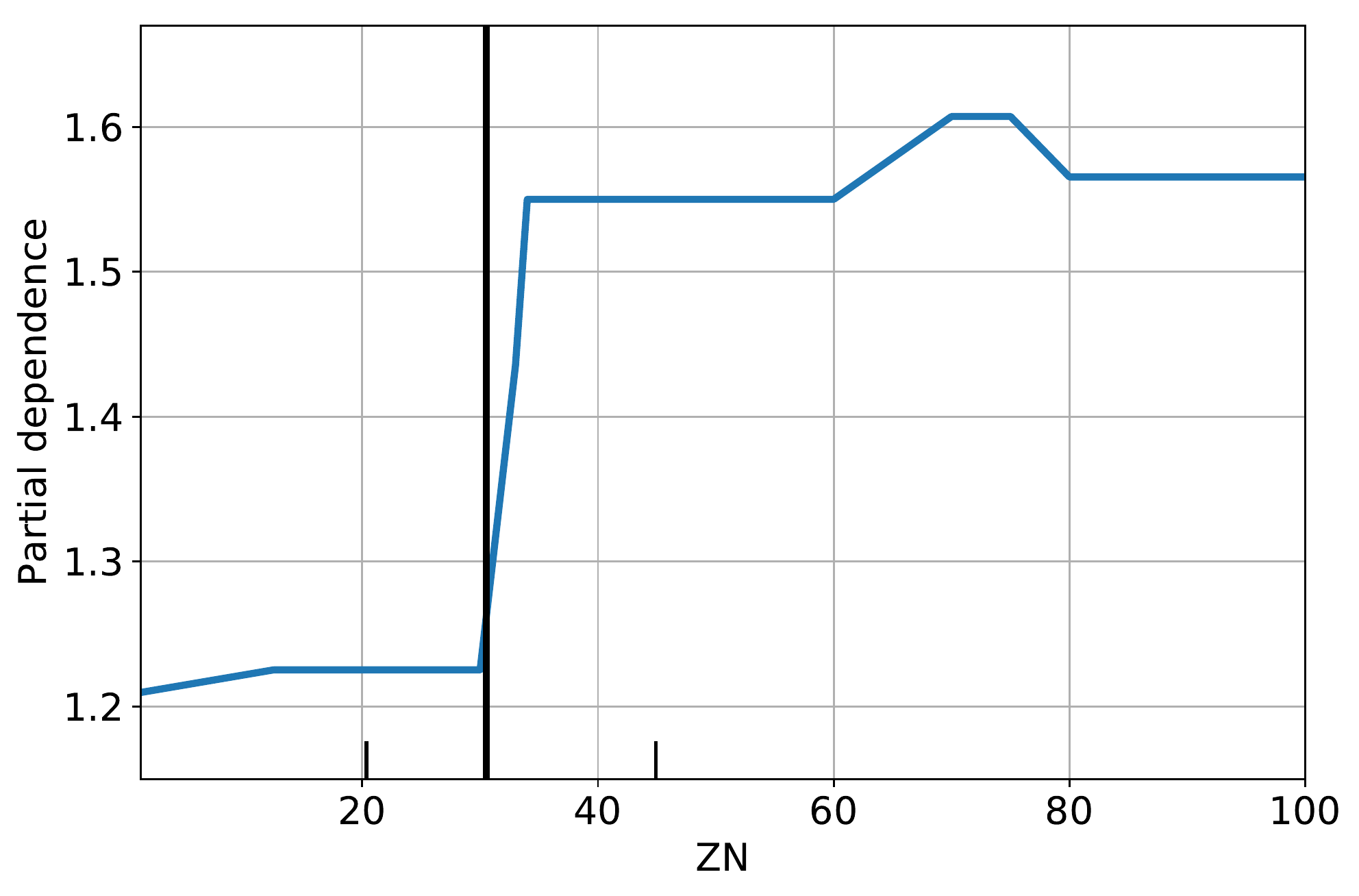}}
\caption{Partial Dependence Plot (PDP) of the gradient boosting model and ZN variable .  Black vertical line indicates variable split generated with SAFE ML method.}
\label{boston_pdp}
\end{center}
\vskip -0.2in
\end{figure}

Partial Dependence Plot for xgboost model and variable ZN is presented in Figure~\ref{boston_pdp}. Flexible boosting model captured the non-linear relationship between variable ZN and target medv. As a result, SAFE ML method divided ZN variable into two binary features to improve performance of refined model.

\subsection{Benchmark on a Number of Tabular Data Sets}
\label{large_benchmark}

In this section we benchmark the SAFE ML method on a~number of tabular data sets for regression and classification problems. We compare performances of three groups of models: simple models trained without SAFE ML feature transformation, complex surrogate models, and refined interpretable models.

\subsubsection{Benchmark for Classification}
\label{benchmark_classification}

We train classification models on six different data sets. They are two simulated data sets, Titanic from Kaggle, Blood Transfusion Service Center \cite{Yeh2009},
Teaching Assistant Evaluation form UCI Machine Learning Repository \cite{Dua:2017}, and Pima Indian Diabetes \cite{johannes1988using}.
 
We use Accuracy and AUC metrics to evaluate models. Logistic regression and classification trees trained without any feature extraction are baselines.
Complex xgboost models are surrogates required to perform SAFE ML algorithm. Parameters of surrogate models differ between data sets. Refined models are logistic regression models and classification trees. 
To chose best penalty for SAFE ML transformations, for each surrogate model we examined 25 equally spaced penalties in the range from~$0.01$~to~$10$. The criterion was performance of a refined model.

Results of benchmarking are in Table~\ref{table_classification}.
For 22 out of 24 cases, refined model surpasses baseline model. In more than half cases, refined model outperforms baseline and surrogate model.

\begin{table}[tb]
\caption{Performances of models trained on six data sets for classification. Artificial data sets are marked by (A). Headers indicate class of baseline model (BASE.) and refined model (REF.). In each case, surrogate model (SURR.) is~xgboost.}
\label{table_classification}
\vskip 0.15in
\begin{center}
\begin{small}
\begin{sc}
\begin{tabular}{lcccr}
\toprule
\multicolumn{4}{c}{Logistic Regression - AUC} \\
\midrule
Data set & BASE. & SURR. & REF. \\
\midrule
SIMULD1  (A)         &  0.833 & \textbf{0.980} & 0.963 \\
SIMULD2  (A)         &  0.897 & 0.982 & \textbf{0.989} \\
Titanic                  & 0.861 & \textbf{0.896} & 0.870  \\
Blood Transfusion        &  0.670 & \textbf{0.679} & 0.668 \\
Teaching Evaluation      &  0.725 & \textbf{0.838} & 0.821 \\
Pima Indian Diabetes     &  0.814 & 0.822 & \textbf{0.838} \\
\toprule
\multicolumn{4}{c}{Logistic Regression - Accuracy}\\
\midrule
Data set & BASE. & SURR. & REF. \\
\midrule
SIMULD1  (A)          &  0.744 & 0.888 & \textbf{0.912} \\
SIMULD2  (A)          &  0.736 & 0.960 & \textbf{0.976} \\
Titanic                   & 0.798 & \textbf{0.834} & \textbf{0.834}  \\
Blood Transfusion         &  0.749 & \textbf{0.754} & 0.668 \\
Teaching Evaluation       &  0.842 & 0.842 & \textbf{0.868} \\
Pima Indian Diabetes      &  0.745 & 0.734 & \textbf{0.771} \\
\toprule
\multicolumn{4}{c}{Classification Tree - AUC} \\
\midrule
Data set & BASE. & SURR. & REF. \\
\midrule
SIMULD1  (A)            &  0.877 & \textbf{0.980} & 0.972 \\
SIMULD2  (A)            &  0.928 & 0 982 & \textbf{0.983} \\
Titanic                     & 0.777 & \textbf{0.896} & 0.878 \\
Blood Transfusion           &  0.598 & 0.667 & \textbf{0.683} \\
Teaching Evaluation         &  0.763 & 0.817 & \textbf{0.842} \\
Pima Indian Diabetes        &  0.665 & \textbf{0.822} & 0.767 \\
\toprule
\multicolumn{4}{c}{Classification Tree - Accuracy}\\
\midrule
Data set & BASE. & SURR. & REF. \\
\midrule
SIMULD1  (A)           &  0.896 & 0.888 & \textbf{0.912} \\
SIMULD2  (A)           &  0.928 & 0.96 & \textbf{0.976} \\
Titanic                    & 0.794 & 0.834 & \textbf{0.839}  \\
Blood Transfusion          &  0.738 & \textbf{0.775} & 0.759 \\
Teaching Evaluation        &  0.842 & 0.842 & \textbf{0.895}  \\
Pima Indian Diabetes       &  0.688 & 0.734 & \textbf{0.760} \\
\bottomrule
\end{tabular}
\end{sc}
\end{small}
\end{center}
\vskip -0.1in
\end{table}

\subsubsection{Benchmark for regression}

In this section, we examine performance of the SAFE ML method on 5 data sets for regression problems. They are Energy Efficiency and Yacht Hydrodynamics form UCI Machine Learning Repository \cite{Dua:2017}, Boston Housing \cite{HARRISON197881}, Warsaw Apartments \cite{DALEX}, and Real Estates \cite{Yeh:2018:BRE:3198938.3199153}.

Base and refined models are linear regression models. We use xgboosts as a surrogate model, xgboost parameters differ between data sets.
To chose best penalty for SAFE ML transformations we examine 25 equally spaced penalties in the range from~$0.01$~to~$10$ and MSE criterion.

Results are presented in Table~\ref{table_regression}. 
For all data sets, baseline models outperform base models. For 3 out of 5 data sets, refined linear model achieves better performance than xgboost model.

\begin{table}[h!tb]
\caption{Performances of models trained on five data sets for regression problem. Artificial data sets are marked by (A). Baseline models (BASE.) and refined models (REF.) are linear regression models. Surrogate models (SURR.) are xgboost models. Performance metric is MSE, values in columns are scaled to MSE for baseline model.}
\label{table_regression}
\vskip 0.15in
\begin{center}
\begin{small}
\begin{sc}
\begin{tabular}{lcccr}
\toprule
Data set & BASE. & SURR. & REF.  \\
\midrule
Warsaw Apartments (A)  & 1 & 7.12 & \textbf{64.99} \\
Real Estates   & 1 & 1.02 & \textbf{1.38}\\
Boston Housing         & 1 & 1.27 & \textbf{1.32}  \\
Energy efficiency     & 1 & \textbf{43.09} & 8.88 \\
Yacht Hydrodynamics   & 1 & \textbf{267.75} & 105.17 \\
\bottomrule
\end{tabular}
\end{sc}
\end{small}
\end{center}
\vskip -0.1in
\end{table}

\subsection{Benchmark Summary}
\label{application_summary}

\mbox{We examined 6 data sets for regression and 5 data sets for classification.}
In more than half of the cases, refined model had outperformed surrogate model. In majority of the rest examples performance differences between surrogate and refined models were minimal.

Refined models are simple, with a small number of parameters, therefore one could conclude that refined models generalize data better than complex models. However, it is worth noting that the refined models generalize relationships that were captured by surrogate models.
Thus, without a complex model as a surrogate, it would not have been possible.
With SAFE ML method, transferring knowledge about relationships to a simple model is automatic and do not require detailed investigation of the complex model.

Even if black-box model gains better results, it is still worth considering applying transparent glass-box model. As we have seen in previous examples, performance of surrogate and refined model were, in general, close to each other. The advantage of a simpler model is that we gain transparency, interpretability and auditability.

\section{Future extensions of the SAFE ML method}
\label{extension}

\subsection{Instance Level Problems}

In previous sections, we showed how to use complex surrogate models to extract global, interpretable features. SAFE ML method could be also extended to instance level feature extraction. A complex model can capture local relationships between variables. Therefore, we may consider several local, interpretable models, instead of one global model.

There are several approaches to obtain locality, we can subset data set, reweight original data, or simulate instances from the original data distribution. One of the examples of local model approximations is LIME (Local interpretable model-agnostic explanations) \cite{lime}. It is a~method for generating local models that approximate the predictions of the underlying complex model. Local models are simple, such as, LASSO regression. Since LASSO is a~method for selecting variables, while applying LIME we perform also a feature extraction.
However, this method is not capable of extracting new interpretable features.

An extension of the LIME that includes extraction of interpretable features is localModel (Local Explanations of Machine Learning Models for Tabular Data) \cite{localModel}. Local interpretable features are created by discretization of numerical features due to the splits of the decision tree. Categorical features are discretized by merging levels using the marginal relationship between the feature and the model response. 
Locality is obtained by generating a~random number of interpretable inputs around the explained instance. Then, LASSO regression model is fitted to new features and original model's responses.

The idea behind localModel is similiar to SAFE ML. However, localModel is used to make statements about predictions and behaviour of the underlying black-box model, while the idea of SAFE is to create new refined model to make its own predictions.

\subsection{Interactions Extractions}
\label{interactions}

SAFE ML algorithm is used for transforming single features. One can consider extending this approach of interactions.
There are methods of capturing interactions from random forest \cite{randomForestExplainer} or xgboost \cite{xgboostExplainer}.
This can be used for extraction of new features which contain information about interactions between variables.

\section{Discussion}
\label{discussion}

In this article, we presented SAFE ML algorithm that uses surrogate model to feature transformations. New features are then used to train refined glass-box model. 

We benchmarked SAFE ML for regression and classification problems. The results confirmed that SAFE ML algorithm produces features that can be further used to fit accurate and transparent model. We also justified the advantage of refined models over surrogate black-boxes.

We also discussed possible extensions of SAFE ML to instance level problems. In addition, we see the possibility of extending the SAFE ML method to include interaction extraction.

\subsection{Benchmarking Methodology}

Benchmarks in Section~\ref{SAFE_application} were based on a single split into training and test data sets. In further research, benchmarks could include k-fold cross-validation technique. However, while applying cross-validation, it would be necessary to take into account values of penalty. In Section~\ref{SAFE_application} we were selecting a penalty on the basis of the model performance on a~test data. The use of cross-validation will cause that values of penalty for each fold will be different. Thus it will not be possible to point the best penalty.

\subsection{Conclusions}
\label{conclusions}

The SAFE ML method allows us to fulfill four requirements of trustworthy predictive model, stated in Section~\ref{motivation}. One can choose a final refined model, accordingly to the simplicity and transparency, therefore statement (3) about interpretability is accomplished. Simple models, such as, linear regression and logistic regression are extensively described from a mathematical point of view. As a result, there are many methods to diagnose such models. Therefore, requirement of the (2) auditability is also fulfilled. In Section~\ref{SAFE_application} we showed that performances of refined models are close to performance of complex surrogate models. Therefore, SAFE ML method allows to gain (1) high model performance. 
In Section~\ref{SAFE_application} we also argued that SAFE ML algorithm allows automatic feature transformation for the purpose of fitting refined model. This approach allows you to omit examining a~complex model. Thus (4) automaticity is also accomplished.

\subsection{Similar Nomenclature}

The phrase \textit{surrogate model} is occasionally referred to an interpretable glass-box model that approximates predictions of a black-box model \cite{h2o_mli_booklet}. The surrogate model in this sense mimics most of the properties of the model under consideration, and is used to makes statements about the black-box model and not about the real world.
However, there is no unambiguous nomenclature for this kind of problem. Models that mimic black-boxes are called also proxy models, shadow models, metamodels, response surface models, emulators \cite{molnar,proc-jsm-2018}.

Therefore, our meaning of the term \textit{surrogate model} is not a duplication the meaning of the existing phrase. In this article, we refer \textit{surrogate model} to a complex model that supports training interpretable model.

\subsection{Software and Code}
\label{software}
Benchmarks from Section~\ref{SAFE_application} were generated with SafeTransformer Python library available at (\url{https://github.com/olagacek/SAFE}). 
Code that generates benchmarks is availible on Github: (\url{https://github.com/agosiewska/SAFE_examples}).

\section{Acknowledgements}

Alicja Gosiewska was financially supported by the grant of Polish Centre for Research and Development POIR.01.01.01-00-0328/17. Przemyslaw Biecek was financially supported by the grant NCN Opus grant 2017/27/B/ST6/01307.

\bibliographystyle{unsrt}
\bibliography{safe}

\end{document}